# Rule-based High-Level Coaching for Goal-Conditioned Reinforcement Learning in Search-and-Rescue UAV Missions Under Limited-Simulation Training

Mahya Ramezani[1] and Holger Voos[1]

*Abstract*— This paper presents a hierarchical decision-making framework for unmanned aerial vehicle (UAV) missions motivated by search-and-rescue (SAR) scenarios under limited simulation training. The framework combines a fixed rule-based high-level advisor with an online goal-conditioned low-level reinforcement learning (RL) controller. To stress-test early adaptation, we also consider a strict no-pretraining deployment regime. The high-level advisor is defined offline from a structured task specification and compiled into deterministic rules. It provides interpretable mission- and safety-aware guidance through recommended actions, avoided actions, and regime-dependent arbitration weights. The low-level controller learns online from task-defined dense rewards and reuses experience through a mode-aware prioritized replay mechanism augmented with rule-derived metadata. We evaluate the framework on two tasks: battery-aware multi-goal delivery and moving-target delivery in obstacle-rich environments. Across both tasks, the proposed method improves early safety and sample efficiency primarily by reducing collision terminations, while preserving the ability to adapt online to scenario-specific dynamics.

## I. INTRODUCTION

UAVs offer a flexible platform for search-and-rescue (SAR) missions that require rapid situational response and autonomous operation in uncertain environments [1]. In representative scenarios, a UAV may need to deliver supplies to multiple locations while avoiding obstacles and managing limited battery reserves, or safely approach a moving person under collision-avoidance and bounded-closing-speed constraints [2, 3]. Such tasks require multi-objective decision-making that balances mission completion, safety, and resource efficiency. In disaster environments, these challenges are further complicated by intermittent visibility, unreliable communication, incomplete maps, and limited prior knowledge of the operating conditions [4]. As a result, UAV autonomy for SAR must remain effective under uncertainty while relying on limited and potentially inaccurate environmental information.

Conventional model-based approaches can perform well when system dynamics, maps, and sensing models are sufficiently accurate. Examples include model predictive control (MPC) [5], control barrier function (CBF) safety layers, and planning pipelines that combine global path planning with local obstacle avoidance [6]. However, in disaster environments these assumptions are often violated. As a result, such methods may require repeated retuning of costs, constraints, and safety margins, and their robustness can degrade under sensing uncertainty and distribution shift.

Reinforcement learning (RL) offers an alternative by learning control policies through interaction to maximize long-horizon return [7]. Deep RL has therefore been explored for UAV navigation and collision avoidance in partially known or unknown environments [8].

Despite its flexibility and the high success rate in various aerospace applications such as drone motion control [9] and path planning [10, 11], satellite collision avoidance [12], satellite docking [13, 14], satellite control [15, 16], and failure predictions [17], deep RL typically requires substantial training experience, which is usually collected in simulation rather than during deployment [18]. This creates a practical limitation for SAR applications: high-fidelity simulators are costly to build and may not adequately capture the sensing and disturbance conditions encountered in the field. As a result, the simulation-to-deployment gap can substantially reduce performance. In addition, early exploration may produce unsafe behavior, and safety-oriented RL variants often require extra mechanisms such as runtime shielding or constraint enforcement [19].

HRL studies decision making with temporal abstraction by decomposing behavior into high level and low-level components. In robotics, this decomposition is often used to separate task level choices [20]. Recent UAV studies adopt this structure by learning a high-level planner that selects intermediate waypoints or modes, while a low-level policy handles local motion and sensing constraints. In multi-UAV search settings, hierarchical multi agent designs have been proposed to separate strategic coordination from low level execution [21]. This structure matches SAR missions, which often switch between search, pursuit, avoidance, and delivery behaviors [22]. At the same time, hierarchical formulations do not remove the need for substantial experience collection. Learning can remain sensitive to the quality of subgoals and to the coupling between levels. Empirical analyses of goal conditioned HRL report that subgoal specification and reachability strongly affect convergence and stability. As a result, early-stage reliability can remain a limiting factor, especially when the agent must perform safely before it has accumulated sufficient experience [23].

This work focuses on a limited-simulation training setting for RL [24]. In this setting, offline training is restricted, and the agent must adapt online under a strict interaction budget during deployment [25]. In SAR, this constraint is well aligned with practice. Large scale simulator training may be unavailable or unreliable, and early unsafe exploration can be costly. Consequently, the learner benefits from additional structure

[1] Automation and Robotics Research Group, Interdisciplinary Centre for Security, Reliability and Trust (SnT), University of Luxembourg.
M. Ramezani (corresponding author; e-mail: mahya.ramezani@uni.lu)
H. Voos (e-mail: holger.voos@uni.lu)

that would typically come from an expert designed strategy, demonstrations, or engineered priors.

In limited-simulation settings, the learner can benefit from additional structure that encodes basic safety and mission priors in an explicit and interpretable form [26]. Recent work has demonstrated the efficacy of such approaches. For instance, a guiding network trained offline with expert demonstrations can adaptively inform policy constraints during online learning, a method proven to be near-optimal. Similarly, a model predictive control-based safety guide can refine a base policy's actions to enforce user-provided constraints, ensuring the final policy is safe without altering the underlying learning objective [27]. More recently, Failure-Aware Offline-to-Online Reinforcement Learning (FARL) has shown that pre-training a world-model-based safety critic offline can dramatically reduce real-world failures during online exploration while actually improving task performance [28]. Likewise, hierarchical frameworks such as SURTR demonstrate how LLM-based high-level planners can be synthesized offline to provide interpretable strategic guidance, while separate low-level RL controllers execute motion planning with their own objectives intact [29]. A practical way to incorporate such structure is to define a fixed rule-based high-level advisor offline from the task specification and system interface, then use it to guide online learning without modifying the low-level RL objective. This preserves real-time compatibility and interpretability while reducing early unsafe behavior.

In this paper, we adopt this perspective by using a fixed, interpretable rule-based high-level advisor that is defined offline from a structured description of the agent capabilities, sensing assumptions, mission objectives, and operational constraints. The low-level component is a goal-conditioned RL controller that updates online. It learns from task-defined rewards using advisor-guided action bias with stronger influence in risky states and mode-aware prioritized replay based on rule-derived metadata. The overall objective is to improve early deployment performance under limited interaction budgets while preserving the ability to adapt online to scenario-specific dynamics. The approach is evaluated on two SAR motivated scenarios. The first scenario is battery aware multi goal delivery in which the UAV must visit multiple targets and reach a terminal destination without battery depletion while avoiding obstacles. The second scenario is delivery to a moving person in which the UAV must approach a moving target while respecting collision avoidance. Performance is compared against non-HRL and baselines using success rate, collision rate, task completion time, and cumulative return, with emphasis on early deployment performance.

The contributions of this work are as follows.
- We formulate a hierarchical decision-making framework for UAV missions motivated by SAR under limited simulation training, covering both battery-constrained multi-goal delivery and moving-target delivery tasks.
- We introduce a fixed, interpretable rule-based high-level advisor that provides mission-aware guidance through recommended actions, avoided actions, and regime-dependent arbitration weights, while remaining compatible with real-time deployment.
- We develop a goal-conditioned low-level RL controller augmented with advisor-guided action bias and mode-aware prioritized replay and show that this design improves early safety and sample efficiency relative to RL baselines in both no-pretraining and limited-pretraining regimes.

## II. PROBLEM DEFINITION

We consider a UAV operating in a SAR-motivated mission and model the environment as a finite-horizon Markov decision process, defined by the tuple $(S, A, P, r, \gamma, T)$. Here $s_t \in S$ denotes the state at time step $t$, $a_t \in A$ denotes the action, $P(s_{t+1} \mid s_t, a_t)$ is the transition kernel, $r(s_t, a_t)$ is the reward, $\gamma \in (0,1)$ is the discount factor, and $T$ is the episode horizon. The objective is to learn a policy $\pi$ that maximizes the expected discounted return.

### A. State and action spaces

We consider planar motion at a fixed altitude $h$. The UAV state is represented in the horizontal plane, and altitude is treated as constant. At time $t$, the agent receives a state that includes UAV kinematics, battery, LiDAR-derived obstacle information, and target related position.

$$s_t = (p_t, \psi_t, v_t, b_t, \ell_t), \quad (1)$$

where $p_t$ is the UAV position, $\psi_t$ is yaw (heading) in the horizontal plane, $v_t$ is the UAV speed, $b_t$ is the battery state of charge, and $\ell_t$ is the LiDAR range vector.

Obstacle perception is provided through a planar LiDAR operating in the horizontal plane. In simulation, LiDAR ranges are generated using a ray-casting model in MATLAB. The environment is represented as an occupancy grid with axis-aligned $1 \times 1$ obstacles. At each time step, we cast $N_\theta = 36$ rays uniformly over $360°$ from the UAV position and compute, for each bearing $\theta_i$, the distance $r(\theta_i)$ to the first intersection with an occupied cell (or obstacle boundary), capped at the maximum sensor range $R_{max} = 10$. The LiDAR observation is then encoded as the range vector $l_t = [r(\theta_1), \dots, r(\theta_{N_\theta})]$, which summarizes local obstacle proximity and is used throughout the paper.

We consider two action parameterizations depending on the low-level RL algorithm. For value-based methods, we use a discrete action space that combines cardinal motion direction with speed adjustment. $a_t = (d_t, v_t^{cmd})$, where $d_t \in \{up, down, left, right\}$, and $v_t^{cmd} \in \{0,1,2,3,4\}$ is the commanded discrete speed level.

For continuous-control methods, we use $a_t = (v_t, \Delta\psi_t)$, where $v_t \in [0, v_{max}]$ is the commanded speed, and $\Delta\psi_t$ is the commanded yaw change over one time step.

### B. Transition model and battery dynamics

State transitions are generated by the simulator using a UAV motion model and an energy-consumption model implemented in the simulator. The simulator propagates position and yaw according to the commanded motion and updates the battery state as a function of the current state and executed action. Battery evolution is modeled as:

$$b_{(t+1)} = \text{clip}(b_t - \Delta t C(s_t, a_t), 0, b_{max}), \quad (2)$$

where $C$ denotes the energy consumption rate induced by the simulator model and

$$\text{clip}(x, 0, b_{max}) = \min(\max(x, 0), b_{max}). \quad (3)$$

The transition kernel $P$ is not provided to the agent, and the policy is learned in a model-free manner from observed transitions.

## C. Tasks and objectives

We study two mission scenarios.

**1) Battery aware multi goal delivery:** The UAV must visit a set of delivery goals $G = g_1, ..., g_K$ and then reach a terminal destination $g_f$. A delivery goal is considered reached when $\| p_t - g_k \| \leq d_{cap}$. The episode is successful if all goals are visited and the terminal destination is reached before battery depletion and without collision.

**2) Moving target delivery:** The UAV must approach a moving target and satisfy delivery conditions defined as $d_g(s_t, g_t) \leq d_{cap}$. In addition to collision avoidance, the task penalizes excessive speed near the target to encourage safe approach behavior.

A collision with an obstacle terminates the episode and incurs a large penalty. Battery depletion, defined by $b_t \leq 0$, also terminates the episode and incurs a large penalty. Successful mission completion yields a positive terminal reward.

## III. PROPOSED METHOD

### A. Overview

We propose an HRL framework for UAV decision-making in SAR missions under limited simulation training. The method combines an offline, rule-based high-level advisor with an online, goal-conditioned low-level controller. The high-level advisor is synthesized from a structured task specification and provides fixed, interpretable guidance that reflects safety and mission priorities. The low-level controller remains responsible for action execution and online adaptation. It operates under limited simulation training and is augmented with advisor-guided action bias (from high-level) and mode-aware experience replay to improve early performance and safety.

### B. High-level coaching via an offline rule design

The high-level component provides an offline fixed advisory prior that guides the low-level goal-conditioned RL controller. Its purpose is to reduce early unsafe behavior by producing a state-dependent recommendation that reflects simple safety and mission priors while leaving long-horizon optimization and adaptation to the low-level learner. The finalized rules are compiled into deterministic code and kept fixed during deployment.

Let $s_t \in S$ be the observed state and let $g_t$ denote the current goal position used by the low-level controller. In multi-goal delivery $g_t$ is the active delivery waypoint (and later the terminal destination), while in moving-target delivery $g_t$ is the current target position. The advisor maps the current state to a recommended action set $A_{rec}(s_t) \subseteq A$, an avoided action set $A_{avoid}(s_t) \subseteq A$, and a state-dependent weight vector $\omega(s_t)$ that modulates the relative contribution of recommendation, avoidance, and exploration within the low-level behavior policy. The advisor does not access the environment transition model, does not perform rollouts, and does not estimate reward-to-go.

### C. State-derived features and the finalized advisory rules

From the LiDAR scan we compute the minimum clearance $d_{min}(s_t) = \min_{1 \leq i \leq N_\theta} r(\theta_i)$, and direction of the closest obstacle. Given the active goal $g_t$, we compute the goal bearing $\psi_g(s_t, g_t) = \text{atan2}((g_t)_y - (p_t)_y, (g_t)_x - (p_t)_x)$. The advisor also computes a free-space heading $\psi_{free}$ that selects a locally high-clearance direction and an emergency repulsive heading $\psi_{rep}$ pointing away from the nearest obstacle. These features are used in the deterministic rules below.

**Final advisory rules**

The advisory rule set defines an advised heading $\psi_{adv}(s_t)$ and an advised speed cap $v_{max}^{adv}(s_t)$, from which a recommended action set or region is constructed. The rules are parameterized by clearance thresholds $r_{stop} < r_{avoid} < r_{clear}$, where $r_{stop}$ is an emergency clearance threshold, $r_{avoid}$ is an avoidance threshold below which the advisor prioritizes free-space seeking, and $r_{clear}$ denotes a nominal clearance regime. Also, near-goal threshold denoted by $d_{near}$, and battery thresholds satisfy $b_{warn} > b_{crit}$. Speed caps satisfy

$$v_{stop} \leq v_{avoid} \leq v_{clear} \leq v_{max}, v_{near} \leq v_{clear} \quad (4)$$

All parameters are fixed for all experiments.

**Advised goal selection**

In multi-goal delivery, the advisor selects the active goal $g_t$ presented to the low-level controller. Let $G_{unv}(t)$ denote the set of unvisited delivery goals. A goal is marked visited when $\| p_t - g \| \leq d_{cap}$. The advisor then assigns

$$g_t = \begin{cases} \text{argmin}_{g \in G_{unv}(t)} \| p_t - g \| & G_{unv}(t) \neq \emptyset \\ g_f & G_{unv}(t) = \emptyset \end{cases} \quad (5)$$

In moving-target delivery, $g_t$ is set to the observed target position.

If the stuck detector activates while pursuing a particular goal, the advisor may temporarily reassign the active goal to the next-best unvisited goal (without marking the current one as visited), for a fixed cooldown horizon, and then revert to the nearest-goal rule. This goal-switching is deterministic and uses only current distances.

**Regime selection and advised heading**

In the emergency regime the advised heading points away from the nearest obstacle, in the avoidance regime it maximizes local free space. Otherwise, it points toward the current goal:

$$\psi_{adv}(s_t) = \begin{cases} \psi_{rep}(s_t) & d_{min}(s_t) \leq r_{stop} \\ \psi_{free}(s_t) & r_{stop} < d_{min}(s_t) \leq r_{avoid} \\ \psi_g(s_t, g_t) & d_{min}(s_t) > r_{avoid} \end{cases} \quad (6)$$

**Advised speed cap**

The advised speed cap is a piecewise function of obstacle clearance:

$$v_{max}^{adv}(s_t) = \begin{cases} v_{stop} & d_{min}(s_t) \leq r_{stop} \\ v_{avoid} & r_{stop} < d_{min}(s_t) \leq r_{avoid} \\ v_{clear} & d_{min}(s_t) > r_{avoid} \end{cases} \quad (7)$$

**Conservative approach near the goal**

To encourage safe delivery near a person and to avoid aggressive motion near the goal, the advisor enforces a near-goal speed cap.

**Battery-aware moderation**

The advisor further reduces allowable speed under low battery to increase conservatism when the remaining energy budget is small:

$$\begin{aligned} b_t \leq b_{warn} & \quad v_{max}^{adv}(s_t) = \min(v_{max}^{adv}(s_t), v_{avoid}), \\ b_t \leq b_{crit} & \quad v_{max}^{adv}(s_t) = \min(v_{max}^{adv}(s_t), v_{stop}) \end{aligned} \quad (8)$$

*D. Recommended and avoided actions*

The advisor converts $\psi_{adv}(s_t)$ and $v_{max}^{adv}(s_t)$ into a recommended action $A_{rec}(s_t)$ subset that steers toward the advised heading under the advised speed cap and an avoided set $A_{avoid}(s_t)$ corresponding to actions that would steer into nearby obstacles or violate conservative safety limits. The avoided set is applied as a soft bias rather than a hard constraint.

**State-dependent weight mapping**

The advisor outputs a weight vector $\omega(s_t) = (\omega_{avoid}(s_t), \omega_{rec}(s_t), \omega_{exp}(s_t))$, where $\omega_i(s_t) \in [0,1]$ and $\sum_i \omega_i(s_t) = 1$. These weights determine the relative emphasis on avoidance-biased actions, advisor-recommended actions, and exploration within the low-level behavior policy. A deterministic regime-based mapping is used:

$$\omega(s_t) = \begin{cases} (\omega_{avoid}^{emg}, \omega_{rec}^{emg}, \omega_{exp}^{emg}) & m_t = emg, \\ (\omega_{avoid}^{obs}, \omega_{rec}^{obs}, \omega_{exp}^{obs}) & m_t = obs, \\ (\omega_{avoid}^{nom}, \omega_{rec}^{nom}, \omega_{exp}^{nom}) & m_t = nom, \end{cases} \quad (9)$$

All numerical weight values are treated as fixed hyperparameters, and held constant across runs to avoid tuning bias.

**Advisor-guided action bias and risk-adaptive arbitration**

The low-level controller uses the advisor as an action-selection bias. At time $t$, the advisor outputs a recommended set $A_{rec}(s_t)$, an avoided set $A_{avoid}(s_t)$, and regime-dependent weights $\omega(s_t)$. These define a risk-adaptive arbitration between advisor guidance and exploration.

Let $m_t \in {emg, obs, nom}$ denote the advisor regime. The behavior policy is

$$\pi_b(\cdot | s_t, g_t) = \omega_{rec}(s_t)\pi_{rec}(\cdot | s_t, g_t) + \omega_{avoid}(s_t)\pi_{safe}(\cdot | s_t, g_t) + \omega_{exp}(s_t)\pi_{exp}(\cdot | s_t, g_t) \quad (10)$$

where $\pi_{rec}$ biases actions toward $A_{rec}(s_t)$, $\pi_{safe}$ suppresses actions in $A_{avoid}(s_t)$, and $\pi_{exp}$ provides exploration. Because $\omega(s_t)$ is regime-dependent, arbitration is more safety-biased in emergency/avoidance states and more exploratory in nominal states.

In discrete control, $\pi_{rec}$ restricts candidates to $A_{rec}(s_t)$, while $\pi_{safe}$ masks actions in $A_{avoid}(s_t)$ unless no feasible alternative exists. In continuous control, the learned action $a_t = (v_t, \Delta\psi_t)$ is biased toward the recommended region by clipping speed to $v_{max}^{adv}(s_t)$ and projecting the post-action heading into the tolerated sector around $\psi_{adv}(s_t)$ when the advisor-selected component is active.

*E. Low-level goal-conditioned control*

The low-level controller is responsible for action execution and online adaptation in the proposed method. It is modeled as a goal-conditioned policy $\pi_L(a_t | s_t, g_t)$ trained from interaction to maximize expected discounted return. Conditioning on the goal variable $g_t$ allows a single policy to represent behaviors for multiple delivery locations and for moving-target delivery. In multi-goal delivery, $g_t$ denotes the active delivery waypoint and is advanced deterministically when the UAV enters the corresponding goal region. After all delivery goals are visited, the terminal destination becomes the active goal. In moving-target delivery, $g_t$ is set to the current target position and is updated at each time step from observations. The low-level policy consumes the relative displacement to the active goal as an input feature, which provides translation invariance without requiring target velocity estimation or an explicit target motion model. The low-level controller uses the discrete and continuous action definitions introduced in Section II-A and learns in a model-free manner under unknown transition dynamics.

The goal-conditioned reward $r(s_t, a_t, s_{(t+1)} | g_t)$ is defined as a task-specific dense reward composed of terminal task signals and intermediate components computed from measurable quantities. At each time step, we apply a constant step penalty to encourage shorter completion time, a near-obstacle penalty based on minimum LiDAR clearance, a goal-reaching reward when entering the capture region of the active goal, and an additional per-step penalty for unusually large battery drops. Concretely, we use

$$r_t = -\lambda_t - \lambda_o \max\left(0, r_{safe} - d_{min}(s_t)\right) + \lambda_p \big(d_g(s_t, g_t) - d_g(s_{t+1}, g_{t+1})\big) + \mathbb{I}_{goal}(t)R_{goal} - \lambda_b \max(0, \Delta b_t - \Delta b_{thr}) + r_{term}(t) \quad (11)$$

where $d_{min}(s_t)$ is the minimum LiDAR range over all bearings, $\Delta b_t = b_t - b_{t+1}$ is the battery drop in one step, and $\Delta b_{thr}$ is a threshold above which battery consumption is penalized. The indicator $\mathbb{I}_{goal}(t) = 1$ when the UAV enters the capture region of the current goal. The terminal term is defined as

$$r_{term}(t) = \begin{cases} R_{all} & \text{the mission is completed at } t \\ -R_{col} & \text{a collision occurs at } t \\ -R_{bat} & \text{battery depletion occurs at } t \\ 0 & otherwise \end{cases} \quad (12)$$

For moving-target delivery, we additionally penalize excessive approach speed near the target:

$$r_{near}(t) = \begin{cases} -\lambda_n \max(0, v_t - v_{near}) & d_g(s_t, g_t) \leq d_{near} \\ 0 & otherwise. \end{cases} \quad (13)$$

and we use $r_t = r_t + r_{near}(t)$.

The low-level controller is trained off-policy using DQN for discrete actions and SAC for continuous actions, with standard architecture implementations [30]. Training occurs online during deployment without assuming extensive repeated pretraining, and experience reuse is provided via the proposed mode-aware prioritized replay mechanism. To improve sample efficiency, the proposed method uses a mode-aware prioritized replay buffer. Each stored transition $\tau_t = (s_t, g_t, a_t, r_t, s_{t+1}, g_{t+1}, done_t)$ stores advisor metadata (regime $m_t \in {emg, obs, nom}$, $d_{min}(s_t)$, and flags for $a_t \in A_{rec}(s_t)$, $a_t \in A_{avoid}(s_t)$). The low-level learner does not infer rule applicability from raw state similarity. Instead, at each step the fixed advisor is evaluated on the current state (and action), and the resulting regime label and rule-activation indicators are stored as replay metadata for prioritization. Priority is defined as

$$p_i = \left(|\delta_i| + \epsilon_{per}\right)^{\alpha_{per}} w_{mode}(m_i) w_{risk}(i) \quad (14)$$

with sampling $P(i) = \frac{p_i}{\sum_j p_j}$, and standard importance-sampling correction ($\beta_{per}$). This prioritizes informative and safety-critical samples without changing the RL update.

To improve safety and sample efficiency, the low-level controller uses the high-level advisor as a fixed behavioral prior through risk-adaptive arbitration and mode-aware prioritized replay. The advisor biases action selection and replay toward safety-critical transitions, while learning remains driven by observed transitions and rewards.

*F. Hyperparameters*

All reward and advisor hyperparameters were selected on a separate validation set using 500 episodes and then fixed for the remainder of the study. DQN and SAC hyperparameters follow standard defaults and are shared across all compared methods. Table I summarizes the complete set of advisor and reward hyperparameters.

**Table I. Advisor and reward hyperparameters.**

| Hyperparameter | Value | Hyperparameter | Value |
|---|---|---|---|
| **Advisor, control, and regime hyperparameters** | | | |
| $r_{stop}$ | 1.0 | $\Delta\psi_{emg}$ | $\pi/6$ |
| $r_{avoid}$ | 2.0 | Top-K directions | K = 2 |
| $r_{clear}$ | 5.0 | $b_{warn}$ | 0.25 |
| $d_{near}$ | 3.0 | $b_{crit}$ | 0.05 |
| $v_{max}$ | 4.0 | $W_{stuck}$ | 10 |
| $v_{stop}$ | 0.5 | $\epsilon_{prog}$ | 1.0 |
| $v_{avoid}$ | 2.0 | $\delta$ | 0.20 |
| $v_{clear}$ | 4.0 | $T_{cool}$ | 40 steps |
| $v_{near}$ | 1.0 | $(\omega_{avoid}, \omega_{rec}, \omega_{exp})_{Emergency}$ | (0.55, 0.35, 0.10) |
| $|\Delta\psi|_{max}$ | $\pi/6$ | $(\omega_{avoid}, \omega_{rec}, \omega_{exp})_{Avoidance}$ | (0.35, 0.40, 0.25) |
| $\Delta\psi_{nom}$ | $\pi/18$ | $(\omega_{avoid}, \omega_{rec}, \omega_{exp})_{Nominal}$ | (0.15, 0.30, 0.55) |
| $\Delta\psi_{avoid}$ | $\pi/10$ | | |
| **Reward hyperparameters** | | | |
| $\lambda_t$ | 0.01 | $R_{bat}$ | 200 |
| $r_{safe}$ | 2.0 | $\Delta b_{thr}$ | 0.004 |
| $\lambda_o$ | 0.5 | $\lambda_b$ | 10 |
| $R_{goal}$ | 50 | $\lambda_n$ | 2.0 |
| $R_{all}$ | 200 | $d_{near}$ | 3.0 |
| $R_{col}$ | 200 | $v_{near}$ | 1.0 |

For the low-level learning algorithms, all deep RL methods use discount factor $\gamma = 0.99$, Adam optimizer, batch size 256, replay buffer size $10^6$, gradient clipping at 1.0, and soft target updates with $\tau = 0.005$, with two-layer MLP (256 units per layer, ReLU). DQN uses learning rate $3 \times 10^{-4}$, Huber loss, Double-DQN, and $\epsilon$-greedy exploration decayed linearly from 1.0 to 0.05 over 200,000 steps. SAC uses actor-critic learning rates $3 \times 10^{-4}$, two critics (clipped double-Q), and automatic entropy tuning with target entropy $-2$ for the action. Mode-aware prioritized replay uses $\alpha_{per} = 0.6$, $\beta_{per}$ linearly annealed from 0.4 to 1.0, and $\varepsilon_{per} = 10^{-6}$. Regime-based replay multipliers are set to $w_{mode}(emg) = 1.5$, $w_{mode}(obs) = 1.2$, and $w_{mode}(nom) = 1.0$. A bounded near-risk replay multiplier $w_{risk} \in [1.0, 1.5]$.

## IV. EXPERIMENTS AND RESULTS

*A. Experiment environment*

We evaluate the proposed framework using simulations conducted in MATLAB 2022. The workspace is a square domain of size $100 \times 100$. Obstacles are axis-aligned $1 \times 1$ squares placed on a uniform grid with obstacle occupancy ratio $\rho \in 0.10, 0.20, 0.30$. Episodes terminate upon collision with an obstacle or battery depletion. A timeout termination is used to bound the interaction budget.

We consider two mission scenarios. In battery-constrained multi-goal delivery, the UAV must visit multi delivery goals $g_1, \ldots, g_K$ and then reach a terminal destination $g_f$. A goal is considered reached when the UAV enters a capture region of radius $d_{cap}$ around the goal. The episode is successful if all delivery goals are reached and the terminal destination is reached before battery depletion and without collision. In moving-target delivery, the UAV must approach a moving target and satisfy a delivery condition based on goal proximity while avoiding collisions. The target follows a deterministic motion policy with scheduled speeds $v_t^{tar} \in 1, 2, 3$, and updates synchronously with the UAV at each time step.

For all experiments we use a fixed episode horizon of $T_{max} = 800$ steps. Each run is generated by sampling a random map at the specified obstacle ratio, sampling start and goal locations in free space, and rejecting instances that are infeasible under straight-line connectivity at the grid resolution. Unless stated otherwise, we use K=3 delivery goals plus a terminal destination for multi-goal delivery. All results are computed using five random seeds per method, and reported statistics aggregate over the full set of evaluation runs.

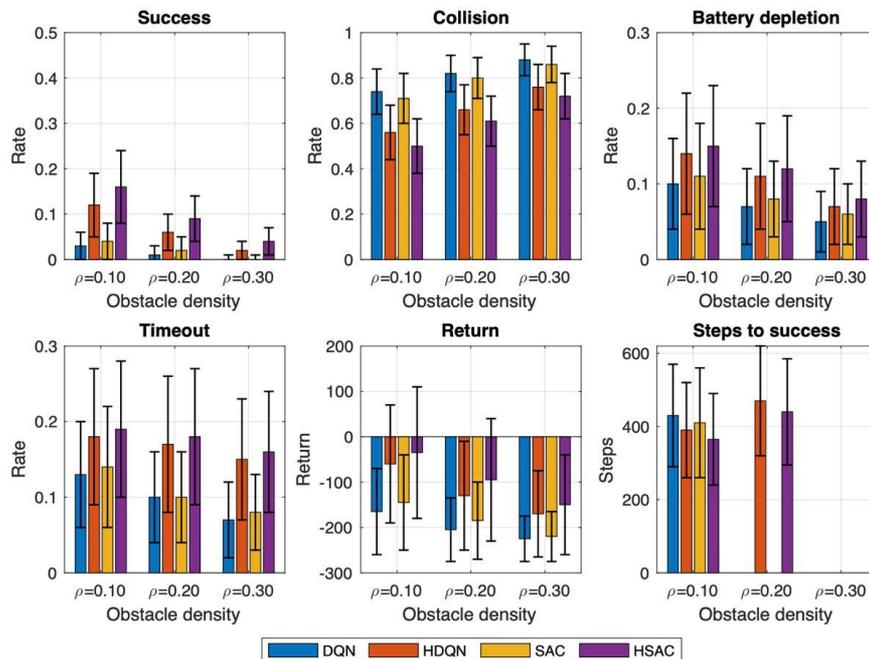

**Fig. 1 The summary of no-pretraining performance for multi-goal delivery across obstacle densities.**

### B. Compared methods

We compare the proposed method against baselines that isolate the effect of hierarchical structure and the advisory prior. The low-level controller is goal-conditioned and trained online using DQN for the discrete action formulation, and SAC for continuous action formulation. For each learning algorithm, we evaluate a baseline without high-level advice and an advised variant that uses the fixed offline advisor (HDQN and HSAC). We additionally report an Advisor-only policy that executes the deterministic advisor without learning. In the no-pretraining moving-target experiments, we compare HSAC to an ablation variant, HSAC (No MA-ER), in which MA-ER is replaced with a standard experience replay buffer while all other components are fixed. This comparison isolates the contribution of mode-aware replay beyond the advisor-guided action bias. All methods share the same state definition, action bounds, termination rules, and reward definition. The advisory rules and hyperparameters are held fixed across environments and run.

### C. Training and evaluation protocol

We report results under three evaluation regimes.

**No-pretraining regime.**

In this setting, the agent is initialized with random network parameters and deployed for a single episode while updating online during that episode. Performance is measured from the deployment episode only. To reduce sensitivity to a particular map instance, results are aggregated over $N_{maps} = 50$ randomly generated maps for each obstacle density, and statistics are reported across five random seeds per method.

**Limited-pretraining regime.**

In this regime, each method is trained for a fixed budget of 1000 (or 2000) episodes sampled from the training distribution. After training, the policy parameters are frozen and evaluated without further learning on a held-out test set of $N_{test} = 200$ maps.

**Test-time generalization regime**.

Finally, we report performance after 5000 training episodes, followed by evaluation in a more challenging, previously unseen environment. Policies are frozen during evaluation. All evaluations use matched random seeds across methods.

### D. Results for the No-Pretraining Online Deployment Regime

We first evaluate the no-pretraining regime, where the policy is deployed with no offline pretraining and is allowed to update only during the deployment episode.

**Battery-Constrained Multi-Goal Delivery**

Fig. 1 summarizes no-pretraining performance for multi-goal delivery (K = 3 plus terminal destination) across obstacle densities. This long-horizon task yields low success for all methods, particularly at moderate and high obstacle density. Advisor-augmented variants (HDQN and HSAC) consistently reduce collisions and improve return, indicating safer and more task-consistent early behavior. In several settings, lower collision rates are accompanied by higher battery depletion or timeout rates, reflecting longer survival without completing the full mission under energy and horizon constraints.

At $\rho = 0.10$, advisor guidance reduces early collisions and improves success, completion time, and return for both discrete and continuous learners. At $\rho = 0.20$, collision reduction remains the primary benefit, while some failures shift to timeout or battery depletion as agents survive longer. At $\rho = 0.30$, successful completion is rare for all methods, nevertheless, HDQN and HSAC maintain lower collision rates and higher returns, indicating that the advisory prior remains beneficial for local safety and feasibility under severe clutter.

**Moving-target delivery**

Tables II and III report performance for moving-target delivery under the no-pretraining regime at $\rho = 0.20$, aggregated over 50 randomly generated maps and five random seeds per method. In contrast to multi-goal delivery, which requires long-horizon route sequencing, moving-target delivery is primarily governed by local reactive interception and collision avoidance. As a result, measurable performance is attainable even in the absence of offline pretraining.

Table II. Moving-target delivery, no-pretraining regime, ρ=0.20.

| Method | Success | Collision | Battery depletion | Timeout | Time to success | Return |
|---|---|---|---|---|---|---|
| SAC | 0.15 ± 0.07 | 0.69 ± 0.11 | 0.02 ± 0.02 | 0.14 ± 0.06 | 230 ± 75 | −120 ± 115 |
| HSAC | 0.29 ± 0.09 | 0.49 ± 0.12 | 0.03 ± 0.02 | 0.19 ± 0.08 | 205 ± 65 | −35 ± 140 |
| Advisor-only | 0.20 ± 0.08 | 0.56 ± 0.11 | 0.03 ± 0.02 | 0.21 ± 0.08 | 215 ± 70 | −65 ± 125 |
| HSAC (No Ma-er) | 0.24 ± 0.08 | 0.55 ± 0.12 | 0.03 ± 0.02 | 0.18 ± 0.08 | 212 ± 70 | −55 ± 135 |

As shown in Table II, the advisor-guided variant (HSAC) demonstrates clear improvements over the baseline SAC controller. Specifically, HSAC increases success from 0.15 to 0.29 while reducing collision rate from 0.69 to 0.49, accompanied by improved time-to-success and higher return. These improvements indicate that the advisory prior promotes safer and more effective early steering decisions during deployment. The Advisor-only policy achieves moderate performance, which is expected in a strict no-pretraining setting where online learning is limited to a single episode and may not fully surpass a well-structured deterministic prior.

The ablation study further clarifies the contribution of mode-aware experience replay. Replacing MA-ER with a standard replay buffer (HSAC No MA-ER) reduces success and increases collision rate relative to full HSAC, demonstrating that mode-aware replay contributes additional gains beyond advisor-guided action arbitration.

Table III. Moving-target delivery, no-pretraining regime, ρ=0.20, by target speed.

| $V_{tar}$ | Method | Success | Collision | Battery depletion | Timeout | Steps to success | Return |
|---|---|---|---|---|---|---|---|
| 1 | SAC | 0.24 ± 0.08 | 0.56 ± 0.12 | 0.03 ± 0.02 | 0.17 ± 0.07 | 210 ± 65 | -60 ± 120 |
| 1 | HSAC | 0.42 ± 0.10 | 0.36 ± 0.11 | 0.04 ± 0.03 | 0.18 ± 0.08 | 185 ± 55 | 40 ± 145 |
| 2 | SAC | 0.15 ± 0.07 | 0.69 ± 0.11 | 0.02 ± 0.02 | 0.14 ± 0.06 | 240 ± 75 | -130 ± 110 |
| 2 | HSAC | 0.30 ± 0.09 | 0.48 ± 0.12 | 0.03 ± 0.02 | 0.19 ± 0.08 | 215 ± 65 | -30 ± 140 |
| 3 | SAC | 0.06 ± 0.04 | 0.82 ± 0.09 | 0.01 ± 0.01 | 0.11 ± 0.05 | 300 ± 95 | -200 ± 80 |
| 3 | HSAC | 0.15 ± 0.06 | 0.63 ± 0.11 | 0.02 ± 0.02 | 0.20 ± 0.08 | 270 ± 85 | -110 ± 125 |

Performance conditioned on target speed is reported in Table III. As expected, increasing speed reduces success and increases collision rates for all methods. Nevertheless, HSAC consistently outperforms SAC across all speeds. At $v_{tar} = 1$, HSAC substantially improves success and reduces collisions. Even at $v_{tar} = 3$, where interception becomes more difficult, HSAC maintains a clear advantage.

*E. Test-set performance after limited pretraining*

We next evaluate test-set performance after fixed offline training budgets of 1000 and 2000 episodes. This regime highlights sample efficiency, since policies are trained on the same distribution and then evaluated without further learning on a held-out test set. Across both tasks, the advisor-augmented variants yield a stronger policy at low training budgets, while the baselines partially close the gap as more experience is collected.

**Table IV. Multi-goal delivery, test performance at ρ=0.20.**

| Episodes | Method | Success | Collision | Battery depletion | Timeout | Steps to success | Return |
|---|---|---|---|---|---|---|---|
| 1000 | DQN | 0.62 ± 0.05 | 0.28 ± 0.07 | 0.06 ± 0.07 | 0.04 ± 0.03 | 260 ± 60 | 250 ± 60 |
| 1000 | HDQN | 0.74 ± 0.09 | 0.18 ± 0.06 | 0.05 ± 0.01 | 0.03 ± 0.02 | 240 ± 55 | 275 ± 55 |
| 2000 | DQN | 0.78 ± 0.06 | 0.16 ± 0.05 | 0.04 ± 0.03 | 0.02 ± 0.02 | 235 ± 55 | 285 ± 50 |
| 2000 | HDQN | 0.87 ± 0.06 | 0.10 ± 0.04 | 0.03 ± 0.02 | 0.02 ± 0.02 | 225 ± 50 | 292 ± 45 |

Results in Table IV show that HDQN achieves higher success and lower collision rates than DQN at both budgets. After 1000 episodes, HDQN improves success from 0.62 to 0.74 and reduces collisions from 0.28 to 0.18, with faster completion and higher return. After 2000 episodes, the DQN baseline improves substantially. However, HDQN remains more reliable and safer, reaching 0.87 success with 0.10 collision rate, compared to 0.78 success and 0.16 collision rate for DQN, while also requiring fewer steps-to-success and achieving comparable-to-higher return. These results indicate that the advisory prior reduces catastrophic failures early in training and continues to provide a safety-oriented inductive bias as learning progresses.

**Table V. Moving-target delivery, test performance at ρ=0.20, $V_{tar} = 2$.**

| Training episodes | Method | Success | Collision | Battery depletion | Timeout | Steps to success | Return |
|---|---|---|---|---|---|---|---|
| 1000 | SAC | 0.58 ± 0.08 | 0.30 ± 0.08 | 0.08 ± 0.04 | 0.04 ± 0.03 | 240 ± 65 | 235 ± 70 |
| 1000 | HSAC | 0.70 ± 0.07 | 0.20 ± 0.07 | 0.07 ± 0.04 | 0.03 ± 0.02 | 225 ± 60 | 260 ± 65 |
| 2000 | SAC | 0.72 ± 0.06 | 0.18 ± 0.06 | 0.07 ± 0.04 | 0.03 ± 0.02 | 225 ± 60 | 265 ± 60 |
| 2000 | HSAC | 0.82 ± 0.06 | 0.11 ± 0.05 | 0.05 ± 0.03 | 0.02 ± 0.02 | 215 ± 55 | 275 ± 55 |

The corresponding results in Table V show a similar pattern, with a stronger early advantage for the advisor-guided method. After 1000 training episodes, HSAC improves success from 0.58 to 0.70 and reduces collision rate from 0.30 to 0.20, while also improving steps-to-success and return. After 2000 episodes, the SAC baseline continues to improve, but HSAC maintains an advantage, reaching 0.82 success versus 0.72 for SAC and reducing collisions from 0.18 to 0.11. The persistence of this gap at 2000 episodes suggests that advisor-guided arbitration and mode-aware experience replay continue to bias learning toward safer, higher-value trajectories, even once the underlying low-level controller becomes substantially more competent.

*F. Test-set performance after Complete pretraining*

We next evaluate a full-training regime in which each method is trained for 5000 episodes and then evaluated without further learning in a harder, previously unseen environment. During training, all methods are trained in a 100×100 workspace with obstacle density $\rho = 0.20$. For moving-target delivery, the target speed is fixed to 2, while for multi-goal delivery we train with $K = 2$ delivery goals. At test time, policies are frozen and evaluated under a distribution shift, obstacle density increases to $\rho = 0.30$, moving-target speed increases to 3, and multi-goal delivery is tested with the harder setting $K = 3$ plus terminal destination. No parameter updates are performed during testing.

Fig. 2 shows the success ratio over training episodes for all methods. Advisor-guided variants converge faster and more consistently, with lower variance across seeds. The effect is stronger in moving-target delivery, which depends mainly on reactive interception and collision avoidance. In multi-goal delivery, the same trend holds, but the final gap is smaller because long-horizon routing and battery and timeout constraints remain limiting factors.

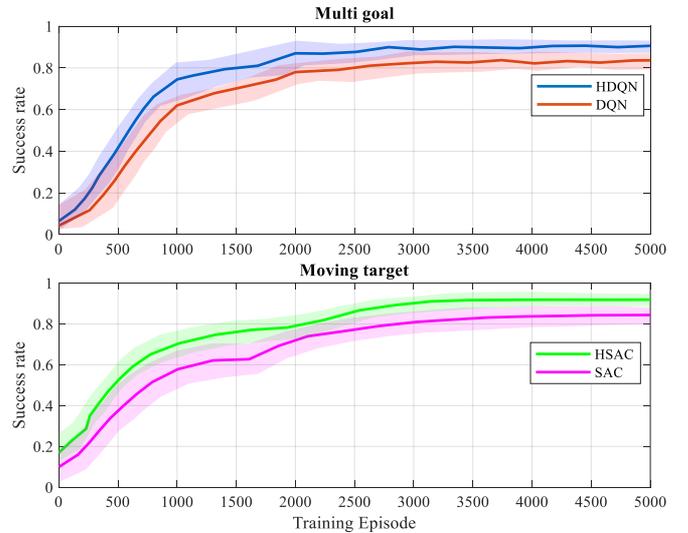

**Fig. 2 The success ratio over training episodes for both Multi goal and Moving target scenarios.**

Results in the test environment are reported in Table VI. In multi-goal delivery, performance degrades substantially for all methods under the combined increase in obstacle density and task complexity, however, advisor-guided methods retain a clear advantage. Relative to DQN, HDQN improves success from 0.20 to 0.32 while reducing collisions from 0.46 to 0.31 and improves efficiency with a markedly higher return. In the continuous-control setting, HSAC achieves the strongest performance, increasing success relative to SAC and reducing collisions, while also completing successful episodes in fewer steps and achieving higher return. Notably, despite improved collision avoidance, battery depletion and timeout rates remain non-negligible across methods, reflecting the fundamental difficulty of completing long-horizon multi-stage missions under energy and horizon constraints in dense clutter.

**Table VI Test performance after complete pretraining in the test environment.**

| Method | Success | Collision | Battery depletion | Timeout | Steps to success | Return |
|---|---|---|---|---|---|---|
| Multi-goal delivery (K = 3 plus terminal). | | | | | | |
| DQN | 0.20 ± 0.06 | 0.46 ± 0.10 | 0.14 ± 0.06 | 0.20 ± 0.07 | 450 ± 110 | -85 ± 115 |
| HDQN | 0.32 ± 0.08 | 0.31 ± 0.09 | 0.16 ± 0.07 | 0.21 ± 0.08 | 395 ± 100 | -5 ± 125 |
| SAC | 0.25 ± 0.07 | 0.40 ± 0.10 | 0.13 ± 0.06 | 0.22 ± 0.08 | 425 ± 105 | -60 ± 120 |
| HSAC | 0.39 ± 0.09 | 0.26 ± 0.08 | 0.15 ± 0.07 | 0.20 ± 0.08 | 365 ± 92 | 35 ± 130 |

| | Moving-target delivery ($v_{tar} = 3$) | | | | | |
|---|---|---|---|---|---|---|
| SAC | 0.44 ± 0.09 | 0.31 ± 0.09 | 0.04 ± 0.03 | 0.21 ± 0.07 | 218 ± 55 | 85 ± 135 |
| HSAC | 0.61 ± 0.08 | 0.18 ± 0.07 | 0.05 ± 0.03 | 0.16 ± 0.06 | 188 ± 47 | 170 ± 120 |
| HSAC No Ma-er | 0.55 ± 0.08 | 0.21 ± 0.07 | 0.05 ± 0.03 | 0.19 ± 0.06 | 195 ± 50 | 145 ± 125 |

In moving-target delivery under faster target dynamics, the advisor-guided methods again generalize more effectively. HSAC improves success over SAC while substantially reducing collision rate and achieves faster completion with higher return. Finally, HSAC (No MA-ER) underperforms full HSAC, supporting the conclusion that mode-aware replay contributes additional robustness beyond advisor-guided arbitration.

## V. CONCLUSION

We presented an advisor-guided hierarchical RL framework for UAV navigation in cluttered environments. By integrating a fixed safety-oriented advisor with online low-level learning and mode-aware replay, the proposed methods improve sample efficiency and reliability.

Across no-pretraining, limited-pretraining, and shifted test-time evaluation, advisor-guided variants consistently reduce collisions and increase success, with the largest gains in moving-target delivery and persistent benefits under distribution shift. In multi-goal delivery, improvements are smaller but remain consistent with safety-driven behavior under battery and horizon constraints.